%% file: main.tex
\documentclass[10pt,twocolumn,letterpaper]{article}

\usepackage[pagenumbers]{cvpr}      
\usepackage{multirow}
\usepackage{booktabs}
\usepackage{float}
\usepackage{tabularx} 
\usepackage{graphicx} 
\usepackage{multicol}
\usepackage{colortbl} 
\usepackage{booktabs} 
\usepackage{pifont}
\newcommand\blfootnote[1]{%
  \begingroup
  \renewcommand\thefootnote{}\footnote{#1}%
  \addtocounter{footnote}{-1}%
  \endgroup
}


\definecolor{cvprblue}{rgb}{0.21,0.49,0.74}
\definecolor{ForestGreen}{rgb}{0, 0.69, 0.31}
\definecolor{00purp}{RGB}{114, 58, 107}
\usepackage[pagebackref,breaklinks,colorlinks,allcolors=cvprblue]{hyperref}

\def\confName{CVPR}
\def\confYear{2025}

\newcommand{\methodname}{\textbf{\textcolor{00purp}{X-Prompt}}\xspace}
\newcommand{\hgreen}[1]{\textcolor{ForestGreen}{\textbf{#1}}} 
\title{\textbf{\textcolor{00purp}{X-Prompt}}: Towards Universal In-Context Image Generation in Auto-Regressive Vision Language Foundation Models}
\vspace{-2mm}
\author{
Zeyi Sun$^{1,2}$, Ziyang Chu$^{2,3}$, Pan Zhang$^{2}$, Tong Wu$^{4}$, Yuhang Zang$^{2}$, \\ Xiaoyi Dong$^{2}$, Yuanjun Xiong$^{6}$, Dahua Lin$^{2,4,5}$, Jiaqi Wang$^{\dagger2}$\vspace{2mm}\\
$^1$Shanghai Jiao Tong University \quad 
$^2$Shanghai AI Laboratory \quad
$^3$Tsinghua University \quad  \\
$^4$The Chinese University of Hong Kong \quad
$^5$CPII under InnoHK \quad
$^6$MThreads AI. \\
\vspace{-2mm}
}

\begin{document}
\maketitle

\blfootnote{$\dagger$ Corresponding author.}
\input{sec/0_abstract}    
\input{sec/1_intro}

\input{sec/2_related_work}
\input{sec/3_method}
\input{sec/4_experiments}
\input{sec/5_discussion}

{
    \small
    \bibliographystyle{ieeenat_fullname}
    \bibliography{main}
}

\input{sec/X_suppl}

\end{document}

%% file: sec/0_abstract.tex
\begin{abstract}

In-context generation is a key component of large language models' (LLMs) open-task generalization capability. By leveraging a few examples as context, LLMs can perform both in-domain and out-of-domain tasks. Recent advancements in auto-regressive vision-language models (VLMs) built upon LLMs have showcased impressive performance in text-to-image generation. However, the potential of in-context learning for general image generation tasks remains largely unexplored. To address this, we introduce \methodname, a purely auto-regressive large-vision language model designed to deliver competitive performance across a wide range of both seen and unseen image generation tasks, all within a unified in-context learning framework. \methodname incorporates a specialized design that efficiently compresses valuable features from in-context examples, supporting longer in-context token sequences and improving its ability to generalize to unseen tasks. A unified training task for both text and image prediction enables \methodname to handle general image generation with enhanced task awareness from in-context examples. Extensive experiments validate the model's performance across diverse seen image generation tasks and its capacity to generalize to previously unseen tasks.

\end{abstract}

%% file: sec/1_intro.tex
\begin{figure*}[t]
\centering
\includegraphics[width=1.0\textwidth]{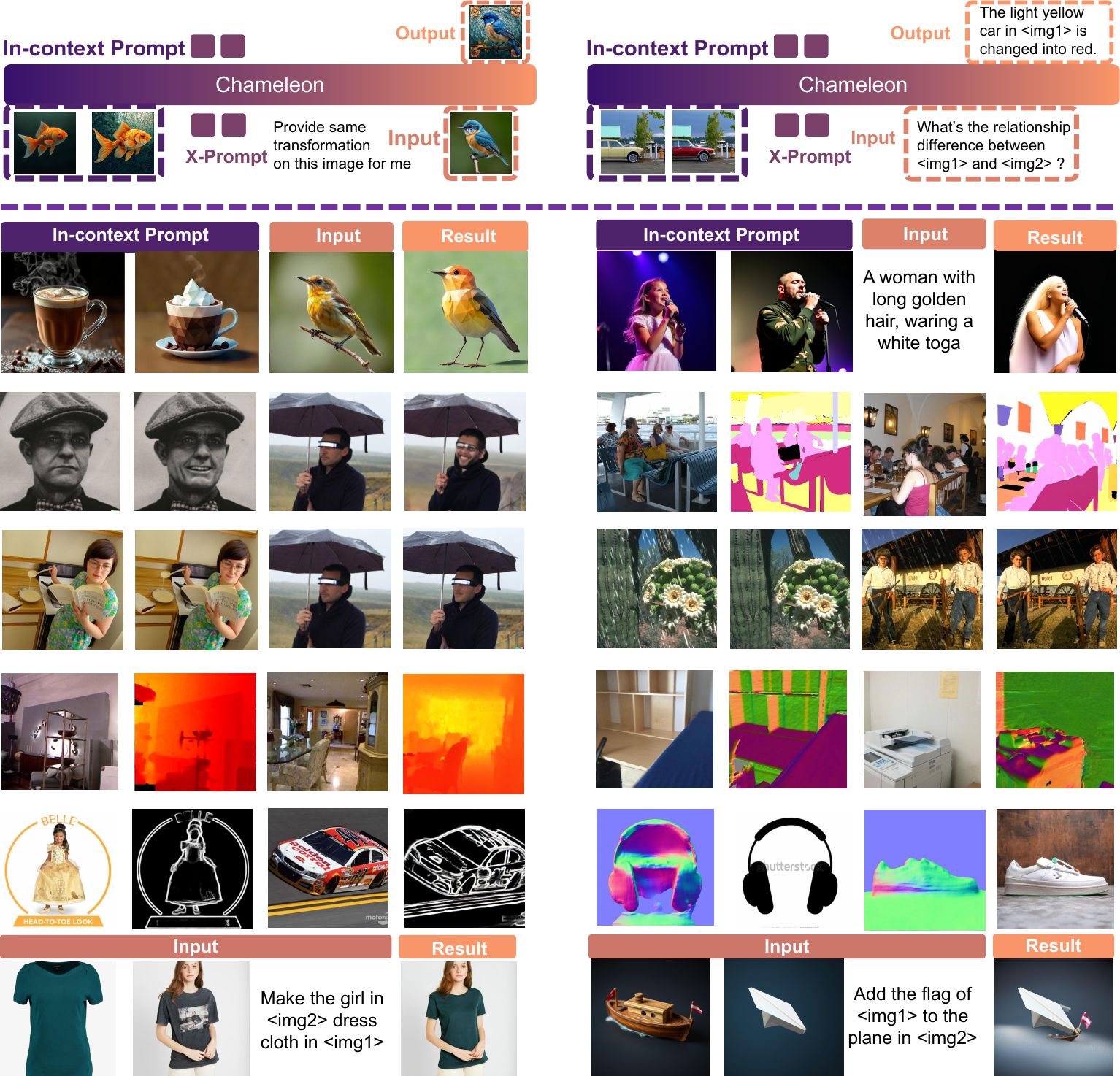}
\captionof{figure}{\textbf{\methodname can perform multi-modal generation based on in-content examples in a pure auto-regressive foundation model.}
}
\label{fig:teaser}
\end{figure*}

\section{Introduction}
Extracting knowledge from a few examples and applying it to novel tasks at inference time has long been a major challenge in machine learning. With the significant success of large language models (LLMs) like GPT-3~\cite{brown2020language} in purely NLP tasks, it has been demonstrated that even a few examples can lead to substantial performance improvements. Previous research has attempted to adapt this capability to computer vision tasks, achieving promising results in vision-only tasks~\cite{wang2023seggpt,wang2023images,gan2023instructcv,bai2024sequential} with pure vision models. However, for tasks that require high-level semantics or text prompt control, like image editing and image personalization, it is important to achieve multi-modal in-context learning.
With the success of multi-modal foundation models, the research focus has shifted to unified multi-modal in-context image generation.

The field of image generation is currently dominated by diffusion models~\cite{song2020denoising,ramesh2021zero,ho2020denoising,liu2024playground,esser2024scaling} like SD~\cite{rombach2022high}, which typically rely on a text encoder alongside a diffusion network. This structure inherently complicates support for in-context learning, as it requires multi-image understanding and reasoning capabilities. Previous approaches~\cite{wu2023next,dong2023dreamllm,sheynin2024emu} like Emu~\cite{sun2023emu} and SEED-X~\cite{ge2024seed}
that bridge diffusion and LLMs often rely on predicted embeddings by LLMs, which introduce huge information loss of image conditions, limiting their abilities to preserve details in editing or low-level vision tasks.
Recently, works like   Chameleon~\cite{team2024chameleon}, Transfusion~\cite{zhou2024transfusion} and concurrent works~\cite{xie2024show,wang2024emu3,xiao2024omnigen,wu2024janus,wu2024vila} have proposed approaches where an LLM directly predicts VQ-VAE~\cite{van2017neural} or VAE~\cite{kingma2013auto} features in an auto-regressive or diffusion manner. 
This approach reduces information loss during image compression, preserving more visual detail while integrating the LLM’s reasoning capabilities. However, limited research has explored in-context image generation on these foundation models.

The primary challenge of in-context learning on these foundation models is the substantial context length required during training. To retain the information in an image, VQ-VAE or VAE image features require a large number of tokens—typically around $(\frac{1}{8})^2$ or $(\frac{1}{16})^2$ of the total image pixels. A single image typically requires 1024–4096 tokens. In an image-to-image in-context task, at least four images are necessary for context, leading to a prohibitive training context length. This limitation, also noted by~\cite{xiao2024omnigen}, restricts support to only three images during training, making direct in-context training impractical due to the excessive context size. 

Moreover, another challenge is improving the ability of foundation models to interpret the intent behind in-context prompts. Since these prompts often consist of several images that implicitly convey the target task without explicit explanations, it is crucial for foundation models to effectively identify and describe the differences or changes between each pair of images.

\begin{figure*}[t]
\centering
\includegraphics[width=1.0\textwidth]{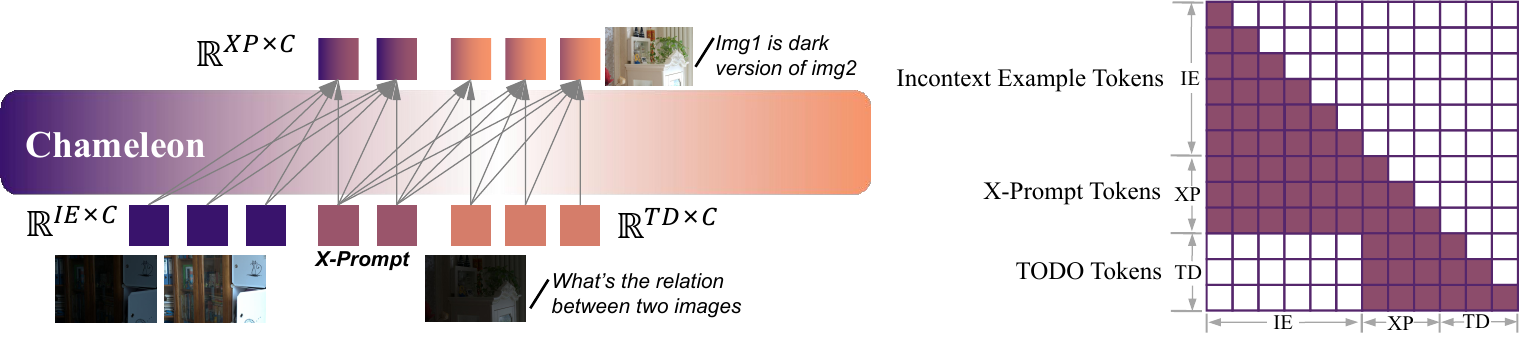}
\caption{\textbf{Attention masking of \methodname for context feature compression and unified text and image next token prediction training.}}
\label{fig:model}
\vspace{-3mm}
\end{figure*}

To address these challenges, we introduce \methodname method to compress the information within in-context examples. Our approach enables the model to distill the content of examples into a sequence of fixed-length compression tokens. During inference, these tokens serve as contextual information for reasoning on new images, effectively reducing the maximum context length required during training. Additionally, compressed tokens of the context enhance the interpretation of target tasks, showing improved generalizability to previously unseen tasks in experiments. Furthermore, unlike pure vision prompt or nature language prompt, Our \methodname supports multi-modal prompt for more diversified tasks like image editing and style-personalization. By constructing reversed tasks and text prediction tasks like generating descriptions of image differences, we enhance the model's overall performance and generalization capabilities. We show some visualization results in \cref{fig:teaser}, where our model can achieve high quality in-context learning on diversified tasks. Our work makes three major contributions:

\begin{itemize}
\item We propose the \methodname method to effectively distill useful information from in-context examples into compressed tokens, improving its performance while reducing the previously prohibitive training context length.
\item By unifying image generation and image description tasks, we significantly enhance Chameleon~\cite{team2024chameleon}’s image generation capabilities.
\item We integrate image generation, editing, dense prediction, and low-level vision tasks into a unified in-context learning framework, demonstrating the effectiveness and generalization of in-context learning across seen and unseen tasks in a pure auto-regressive vision-language foundation model.
\end{itemize}

\label{sec:intro}

%% file: sec/2_related_work.tex
\section{Related Work}

\noindent \textbf{Large Vision-Language Models (LVLMs).}
The emergence of large language models (LLMs)~\cite{Brown2020LanguageMA,openai2023gpt,Chowdhery2022PaLMSL,Anil2023PaLM2T,Hoffmann2022TrainingCL,Touvron2023LLaMAOA} have made remarkable breakthroughs. The domain of research has increasingly turned its attention toward Large Vision-Language Models (LVLMs). Previous advances in this field focus on the integration of vision understanding capabilities with LLMs~\cite{2023GPT4VisionSC,zhang2023internlmxcomposer,Li2023BLIP2BL,Awadalla2023OpenFlamingoAO,dong2024internlmxcomposer2,sun2023alphaclip,qi2023gpt4point,liu2024mia,xing2024pyramiddrop,huang2024deciphering,qi2024gpt4point}. Recent works start to focus on integrating vision generation abilities. One early line of these works~\cite{sun2023emu,wu2023next,ge2023making,ge2024seed,sun2024generative,huang2024smartedit} compress visual features with LLMs into compressed embeddings and use diffusion decoder (like SDXL~\cite{podell2023sdxl}) to generate visual contents, However, this suffers great information loss during LLM encoding process, leading to unsatisfying results in image editing tasks. Another line of work, pioneered by Chameleon~\cite{team2024chameleon}, uses unified image tokens from VQ-VAE~\cite{van2017neural,esser2021taming} to unify perception and generation. In this work, we aim to fully unlock the potential of Chameleon for general image generation in a unified in-context learning paradigm for both seen and unseen tasks.

\noindent \textbf{Auto Regressive based Image Generation.} 
While previous state of the art image generation dominant by diffusion models~\cite{rombach2022high,song2020denoising,ramesh2021zero,ho2020denoising,liu2024playground,esser2024scaling,bu2024broadway,ling2024motionclone,qi2024tailor3d,yang2024layerpano3d,he2024freeedit}, recent works on auto-regressive for image generation have shown promising results~\cite{sun2024autoregressive,zhang2024var,liu2024lumina,wang2024emu3,luo2024open,tian2024visual,ma2024star,wu2024vila}. However, these models only shows experiments results on text-to-image generation~\cite{sun2024autoregressive,wu2024janus,li2024imagefolder,wu2024vila,tang2024hart,wu2024vila}, with limited research on other types of image generation tasks or lack of quantitative results~\cite{yu2023scaling,sun2023emu}. In this work, we not only enhance text-image alignment but also extend our exploration to diverse image generation tasks, including image editing, controlled image generation, and perception tasks such as semantic segmentation and depth estimation. We demonstrate that auto-regressive models can achieve competitive results across these tasks in a unified framework.

\noindent \textbf{In-Context Learning.} GPT-3~\cite{brown2020language} introduced the paradigm of in-context learning, where diverse NLP tasks are reformed as text completion tasks, enhanced through prompts containing embedded examples, which significantly boosts performance on related tasks. In-context learning has also been explored in the vision domain~\cite{bar2022visual,wang2023images,wang2023seggpt,bai2024sequential}, but these models are limited to vision-only tasks, lacking multi-modal versatility. Multi-modal models like Emu-1/2~\cite{sun2023emu,sun2024generative} demonstrate in-context learning capabilities; however, their reliance on embeddings predicted by LLM from image features and the integration of an external diffusion model restrict their effectiveness in image editing and dense prediction tasks. In contrast, our approach utilizes a unified early-fusion representation based on  Chameleon~\cite{team2024chameleon}, enabling a single model to generalize across a broader array of tasks with improved performance and enhanced generalizability.

%% file: sec/3_method.tex
\section{Method}

\begin{figure*}[t]
\centering
\includegraphics[width=1.0\textwidth]{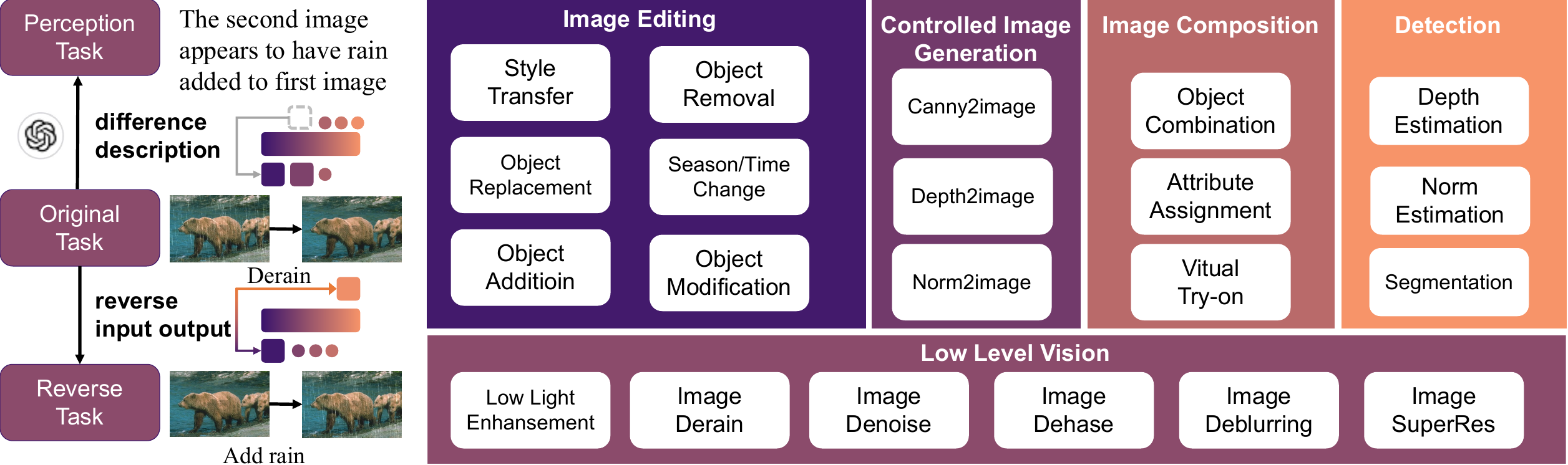}
\caption{\textbf{Training data pair augmentation and list of training prototype tasks and subtasks.} We introduce reverse task and difference description task through next text token prediction to improve the performance and generalizibility.}
\label{fig:data}
\vspace{-2mm}
\end{figure*}

\subsection{In-Context Example Compression}
\label{sec:attn_mask}



\noindent We introduce a context-aware compression mechanism to better model in-context examples within Chameleon. As illustrated in \cref{fig:model}. This mechanism defines three types of tokens: In-Context Example Tokens (IE), X-Prompt Tokens (XP), and TODO Tokens (TD). Given an in-context example as a prompt, Chameleon embeds it into a feature space, producing In-Context Example Tokens \( X_{IE} \in \mathbb{R}^{IE \times C} \), where \( IE \) denotes the number of tokens and \( C \) the feature dimension. The model further includes learnable X-Prompt Tokens, represented as \( X_{XP} \in \mathbb{R}^{S \times C} \), where \( S \) is the number of learnable tokens optimized during training. 

To encourage the model to store contextual information in \( X_{XP} \), we disconnect the relationship between \( X_{IE} \) and the TODO Tokens \( X_{TD} \) through attention masking, forcing the model to rely on X-Prompt Tokens for context representation. The model then generates the TODO Tokens \( X_{TD} \in \mathbb{R}^{TD \times C} \), with \( TD \) as the target sequence length, by sequentially maximizing the conditional probability of each token in \( X_{TD} \) given \( X_{XP} \) and all previously generated tokens \( X_{TD_{<t}} \), as formulated below:

\[
\begin{aligned}
p_\theta(X_{TD} \mid X_{XP}) &= \prod_{t=1}^{TD} p_\theta(X_{TD_t} \mid X_{XP}, X_{TD_{<t}}) \\
&= \prod_{t=1}^{TD} \text{softmax}(f(X_{XP}, X_{TD_{<t}}; \theta))_{X_{TD_t}}
\end{aligned}
\]
where \( X_{TD_t} \) denotes the \( t \)-th token in the TODO sequence, and \( X_{TD_{<t}} \) represents all previously generated TODO tokens with model $f$, parameterized by weights $\theta$. This approach enables Chameleon to effectively compress valuable features from in-context examples, enhancing its capability for context-aware generation and reduce training context length.

\subsection{Task Augmentation Pipeline}
\label{sec:data_aug}

As Chameleon~\cite{team2024chameleon} possess both text and image generation ability, we also adopt unified training on interleaved text and image generation. As illustrated in \cref{fig:data}, to further augments the training data, We construct a data generation pipeline. Each generation task is converted into a text prediction task that focuses on describing the relationship between the input and output images. This text prediction task requires the model to interpret and articulate the relational changes between the images. Advanced vision-language models, such as GPT-4V~\cite{2023GPT4VisionSC} or QwenVL-2~\cite{wang2024qwen2}, are employed for this purpose, as they can generate and interpret open-ended relational descriptions. The detailed prompts are available in \cref{sup:diff_description}. Through training the model to describe differences by generating text tokens, we equip it with a deeper understanding of the relationship of input and output images, which enhances its generalization ability and improves the performance of image generation.

Additionally, we introduce a task-reversion augmentation. For each task, such as ``deraining" (removing rain from an image), we introduce a reverse task—``adding rain"—by swapping the input and output. This strategy effectively doubles the task variety, enabling the model to learn transformations in both directions and deepening its comprehension of the underlying transformation processes.

\begin{table*}[t]
\centering
\setlength\tabcolsep{12pt}
\scalebox{0.8}{
\begin{tabular}{c|lccccccc}
\toprule
Type & Model & Single Obj. & Two Obj. & Counting & Colors & Position & Color Attri. & Overall \\
\midrule
\multirow{7}{*}{Diffusion} 
& LDM~\cite{rombach2022high} & 0.92 & 0.29 & 0.23 & 0.58 & 0.02 & 0.05 & 0.37 \\
& SD-1.5~\cite{rombach2022high} & 0.97 & 0.38 & 0.35 & 0.76 & 0.04 & 0.06 & 0.43 \\
& SD-2.1~\cite{rombach2022high} & 0.98 & 0.51 & 0.44 & 0.85 & 0.07 & 0.17 & 0.50 \\
& DALL-E 2~\cite{ramesh2022hierarchical} & 0.94 & 0.66 & 0.49 & 0.77 & 0.10 & 0.19 & 0.52 \\
& Show-o~\cite{xie2024show} & 0.95 & 0.52 & 0.49 & 0.82 & 0.11 & 0.28 & 0.53 \\
& SDXL~\cite{podell2023sdxl} & 0.98 & 0.74 & 0.39 & 0.85 & 0.15 & 0.23 & 0.55 \\
& DALLE 3~\cite{betker2023improving} & 0.96 & 0.87 & 0.47 & 0.83 & 0.43 & 0.45 & 0.67 \\
\midrule
\multirow{5}{*}{Auto-regressive} 
& LLamaGen~\cite{sun2024autoregressive} & 0.71 & 0.34 & 0.21 & 0.58 & 0.07 & 0.04 & 0.32 \\
& Emu3Gen~\cite{wang2024emu3} & 0.98 & 0.71 & 0.34 & 0.81 & 0.17 & 0.21 & 0.54 \\
& Chameleon~\cite{team2024chameleon} & - & - & - & - & - & - & 0.39 \\
& Ours & 0.97 & 0.69 & 0.28 & 0.71 & 0.14 & 0.15 & 0.49 (\hgreen{+0.10}) \\
& Ours (+text pred.) & 0.98 & 0.73 & 0.33 & 0.85 & 0.26 & 0.28 & 0.57 \\
& $\Delta$ & \hgreen{+0.01} & \hgreen{+0.04} & \hgreen{+0.05} & \hgreen{+0.14} & \hgreen{+0.12} & \hgreen{+0.04} & \hgreen{+0.08} \\
\bottomrule
\end{tabular}
}
\vspace{-3mm}
\caption{\textbf{Evaluation of text-to-image generation ability on GenEval~\cite{ghosh2024geneval} benchmark.} Unifying image dense description task through next text token prediction can significantly improve the text-image alignment of images generated by Chameleon~\cite{team2024chameleon}.}
\vspace{-3mm}
\label{tab:text2image}
\end{table*}

\subsection{Retrieval-Augmented Image Editing}
\label{sec:raie}
Following the spirit of Retrieval-Augmented Generation (RAG)~\cite{lewis2020retrieval}. We introduce Retrieval-Augmented Image Editing (RAIE) to enhances image editing by retrieving the relevant examples from a database. Given an input image \( I_{\text{input}} \) and an instruction \( \text{Instr}_{\text{current}} \), RAIE searches for the most similar instruction \( \text{Instr}_{\text{retrieved}} \) and corresponding image editing pair\( P_{\text{retrieved}} \) in the database. The retrieval process is defined as:

\[
(\text{Instr}_{\text{retrieved}}, P_{\text{retrieved}}) = \underset{(\text{Instr}, P) \in D}{\text{argmin}} \; \text{dist}(\text{Instr}, \text{Instr}_{\text{current}}).
\]
where we use the cosine similarity of CLIP~\cite{radford2021learning} text features as the distance function \( \text{dist} \). This retrieved example serves as in-context guidance for the model, which then generates the edited output \( I_{\text{output}} \) based on both \( \text{Instr}_{\text{current}} \) and the retrieved pair \( (\text{Instr}_{\text{retrieved}}, P_{\text{retrieved}}) \):

\[
I_{\text{output}} = \text{Model}(I_{\text{input}}, \text{Instr}_{\text{current}}, \text{Instr}_{\text{retrieved}}, P_{\text{retrieved}}).
\]
It is worth noticing that \( \text{dist} \) can be extended to other function beyond simply compute text feature similarity. This automated retrieval process reduces the need for manual intervention and can also be customized by users to achieve more precise, tailored image editing. RAIE is an optional choice, and we apply this method only when specifically mentioned in the experiment section.

By leveraging in-context examples, RAIE enhances the consistency and accuracy of image editing tasks. RAIE is naturally suited for a unified auto-regressive model. However, it poses significant challenge for current state-of-the-art diffusion models, which rely solely on text encoders and lack comprehensive understanding capabilities.

%% file: sec/4_experiments.tex
\section{Experiments}
Though we train a unified model across different tasks, we report the data preparation process separately in each subsection for clarity. The complete training dataset comprises approximately 5 million data pairs, expanding to around 8 million pairs after task reversion and text prediction task augmentation (detailed in \cref{sup:data}). For all tasks that take an image with text as input and produce an image with text as output, we include an in-context example of the same task type. We set the batch size to 1024 and the context window size to 5120. The learning rate is set to 1e-4 with a cosine learning rate scheduler. Training is conducted on 128 NVIDIA A100-80G GPUs over 20,000 steps.

\subsection{Text-to-Image Generation}

\noindent \textbf{Settings.} To enhance Chameleon’s~\cite{team2024chameleon} text-to-image generation capabilities, we utilize QWen-VL2~\cite{wang2024qwen2} to rewrite dense descriptive captions for 500K high-quality images filtered from the LAION-Aesthetic~\cite{schuhmann2022laion} dataset (detailed in \cref{sup:cap_laion}), selecting only images with an aesthetic score greater than 6. For evaluation, we adopt the GenEval~\cite{ghosh2024geneval} benchmark.

\noindent \textbf{Results.} As shown in \cref{tab:text2image}, we significantly enhance Chameleon’s original text-to-image generation capabilities. Leveraging the  image dense description task, our model further achieves competitive results compared to other auto-regressive models for text-to-image generation. This experiment highlights the effectiveness of unifying dense image description and image generation tasks, resulting in notable improvements, particularly in tests involving complex multi-object and color attributes. The ability to generate images that accurately follow text prompts is essential for our downstream applications like image-editing. Qualitative visualization are available in \cref{sup:t2i}.

\subsection{Image Dense Prediction}

\begin{table*}[t]
    \centering
    \resizebox{1.0\textwidth}{!}{%
        \begin{tabular}{lrccccccccc}
            \toprule
            \multirow{3}{*}{\textbf{Type}} & \multirow{3}{*}{\textbf{Methods}} & {\bf Depth Est.} & \bf Semantic Seg. & \bf Surface Normal Est. &
            \multicolumn{2}{c}{\bf Lowlight Enhans.} & \multicolumn{2}{c}{\bf Deblur} & \multicolumn{2}{c}{\bf Derain} \\
            
            & & RMSE↓ & mIoU↑ & Mean Angle Error↓ &
            PSNR↑ & SSIM↑ & PSNR↑ & SSIM↑ & PSNR↑ & SSIM↑ \\
            
            \cmidrule(lr){3-3} \cmidrule(lr){4-4} \cmidrule(lr){5-5} 
            \cmidrule(lr){6-7} \cmidrule(lr){8-9} \cmidrule(lr){10-11}
            
            & & {NYUv2} & {ADE20K} & {NYU-Depth V2} &
            \multicolumn{2}{c}{LOL} & \multicolumn{2}{c}{GoPro} &
            \multicolumn{2}{c}{Rain100L} \\
            
            \midrule
            \multirow{7}{*}{Domain Specific Model} &
 DepthAnything~\cite{yang2024depth} & \cellcolor{gray!30} 0.206  \\
            & Marigold~\cite{ke2024repurposing} & \cellcolor{gray!30} 0.224  \\
            & Mask DINO~\cite{li2023mask} & &  \cellcolor{gray!30} 60.80 &  \\
            & Mask2Former~\cite{cheng2022masked} & & \cellcolor{gray!30} 56.10 &  \\
            & Bae et al.~\cite{bae2021estimating} & & & \cellcolor{gray!30} 14.90 \\
            & InvPT~\cite{ye2022inverted} & & & \cellcolor{gray!30} 19.04  \\
            & AirNet~\cite{li2022all} & & & & \cellcolor{gray!30} 18.18 & \cellcolor{gray!30} 0.735 & \cellcolor{gray!30} 24.35 & \cellcolor{gray!30} 0.781 & \cellcolor{gray!30} 32.98 & \cellcolor{gray!30} 0.951 \\
            & InstructIR~\cite{conde2025instructir} & & & & \cellcolor{gray!30} 23.00 & \cellcolor{gray!30} 0.836 & \cellcolor{gray!30} 29.40 & \cellcolor{gray!30} 0.886 & \cellcolor{gray!30} 36.84 & \cellcolor{gray!30} 0.937 \\
            \midrule
            \multirow{4}{*}{Unified Model (continuous)} & Painter~\cite{wang2023images} & 0.288  & 49.90 & $\times$ & 22.40 & 0.872 & $\times$ & $\times$ & 29.87 & 0.882  \\
            & InstructCV~\cite{gan2023instructcv} & 0.297 & 47.23 & $\times$ & $\times$ & $\times$ & $\times$ & $\times$ & $\times$ & $\times$ \\
            & InstructDiffusion~\cite{geng2024instructdiffusion} & $\times$ & $\times$ & $\times$ & $\times$ & $\times$ & 23.58 & - & 19.82 & 0.741 \\
            & OmniGen~\cite{xiao2024omnigen} & 0.480 & $\times$ & $\times$ & 13.38 & 0.392 & 13.39 & 0.321 & 12.02 & 0.233 \\
            \midrule
            \multirow{3}{*}{Unified Model (discrete)} & Unified-IO~\cite{lu2022unified} & 0.387 & 25.71 & - & $\times$ & $\times$ & $\times$ & $\times$ & $\times$ & $\times$ \\
            & Lumina-mGPT~\cite{liu2024lumina} & $\times$ & 20.87 & 22.10 & $\times$ & $\times$ & 17.59 & 0.536 & 16.61 & 0.365 \\
            & Ours & 0.277 & 31.21 & 19.17 & 19.71 & 0.810 & 21.04 & 0.761 & 25.53 & 0.843 \\
            \bottomrule
        \end{tabular}
    }
    \vspace{-2mm}
    \caption{\textbf{Comparison of \methodname with task-specific and vision generalist baselines} across six representative tasks, covering both high-level visual understanding and low-level image processing. '$\times$' indicates that the method is incapable of performing the task.
    }
    \label{tab:dense_predict}
\end{table*}

\begin{figure*}[t]
\centering
\includegraphics[width=1.0\textwidth]{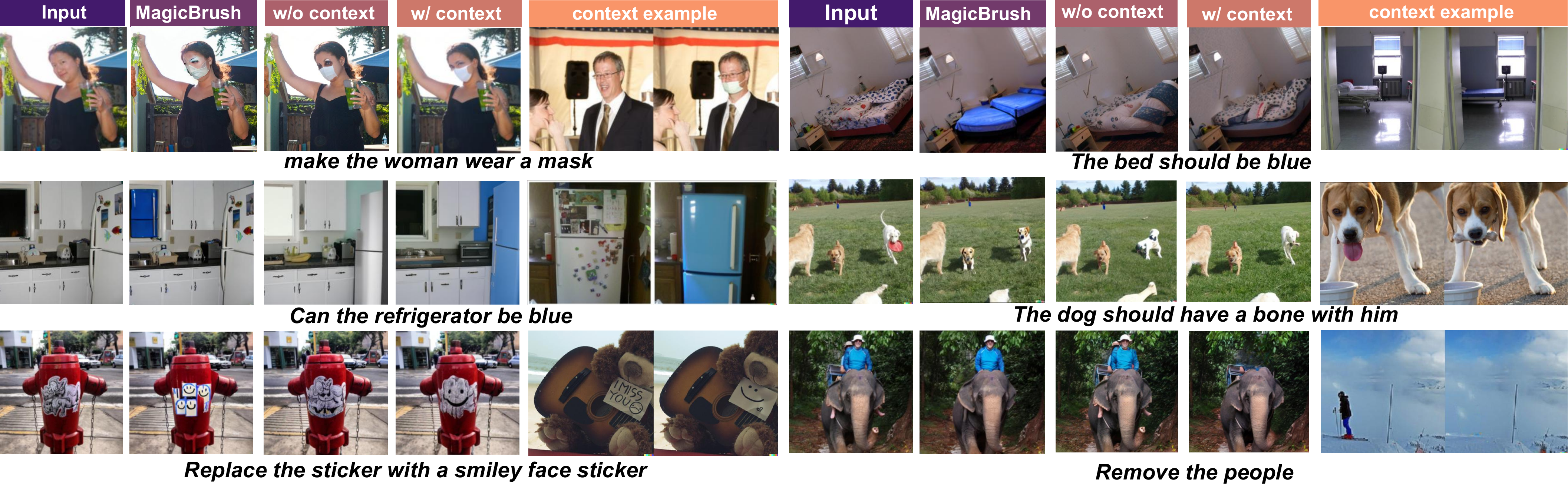}
\vspace{-4mm}
\caption{\textbf{Qualitative Results on MagicBrush~\cite{zhang2024magicbrush} testset} comparing with MagicBrush results w/ and w/o context examples.}
\label{fig:edit}
\vspace{-3mm}
\end{figure*}

\noindent \textbf{Settings.} We use representative datasets for dense prediction tasks: NYU-v2~\cite{silberman2012indoor} for depth and surface normal estimation, ADE-20K~\cite{zhou2017scene} for semantic segmentation, and Rain-13K~\cite{fu2017removing}, LOL~\cite{wei2018deep}, and GoPro~\cite{nah2017deep} for corresponding low-level vision tasks. Full training data details are available in the \cref{sup:data}.

\noindent \textbf{Results.}
 We select typical specialist model and previous vision generalist for comparison. Results are shown in \cref{tab:dense_predict}, where our model can achieve competitive results on dense prediction task and low level vision task. Our model is the first to deliver promising results using a unified, discrete token approach. The slight performance gap on low level vision task compared to continuous feature prediction models is due to the inherent information loss in the VQ-VAE discretization process, as Chameleon~\cite{team2024chameleon} adopt 16x compression rate. However, by discretizing images and adopting a next-token prediction approach akin to that used in large language models, our method offers promising scalability for future advancements.

\subsection{Image Editing with RAIE}

\noindent \textbf{Settings.} 
As we introduced Retrieval-Augmented Image Editing (RAIE) in \cref{sec:raie}, we also prepare training data in the same way. We use publicly available UltraEdit~\cite{zhao2024ultraedit} (500K) and MagicBrush~\cite{zhang2024magicbrush} (8K) For the training. We use CLIP-B/32~\cite{radford2021learning} text encoder to encode the the edit instruction for each training sample and retrieval the most similar instruction feature as its neighbor sample (excluding the sample itself). During training, we input the neighboring sample as a task prompt context and prompt the model to predict the edited image.
Besides generation task, we also use QWen-VL2~\cite{wang2024qwen2} to describe the differences between the input image and the edited image. We add these difference description tasks into the training to help model gain better understanding of images and their variations. Preparing each editing pair with a similar editing pair for in-context learning is crucial to the success of RAIE, as we frequently observe similar editing pairs in both the Ultra-Edit and MagicBrush datasets. We provide detailed analysis of this in \cref{sup:raie}.

For evaluation, We use MagicBrush~\cite{zhang2024magicbrush} benchmark, we also encode the edit instuction using CLIP-B/32 text encoder. We only use MagicBrush training set as reference database to perform Retrieval-Augmented Image Editing (RAIE) proposed in ~\cref{sec:raie}. The testing metics and CLIP and DINO~\cite{caron2021emerging} models are consistent with ~\cite{zhao2024ultraedit,zhang2024magicbrush}.

\begin{table}[ht]
\setlength{\tabcolsep}{4pt}  
\renewcommand{\arraystretch}{1.2}  
\centering
\scalebox{0.72}{
\begin{tabular}{llccccc}
\toprule
Type & Methods & $\text{CLIP}_{\text{dir}}$↑ & $\text{CLIP}_{\text{out}}$↑ & $\text{CLIP}_{\text{img}}$↑ & DINO↑ \\
\midrule
\multirow{3}{*}{Continuous} & InstructPix2Pix~\cite{brooks2023instructpix2pix} & 0.081 & 0.276 & 0.852 & 0.750 \\
& MagicBrush~\cite{zhang2024magicbrush} & 0.106 & 0.278 & 0.933 & 0.899 \\
& UltraEdit~\cite{zhao2024ultraedit} & 0.093 & 0.274 & 0.899 & 0.848 \\
\midrule
\multirow{4}{*}{Discrete} & Lumina-mGPT~\cite{liu2024lumina} & 0.025 & 0.253 & 0.810 & 0.751 \\
& Ours (w/o text pred.) & 0.067 & 0.263 & 0.823 & 0.785 \\
& Ours (w/ text pred.) & 0.083 & 0.271 & 0.857 & 0.781 \\
& Ours + RAIE & 0.097 & 0.279 & 0.862 &  0.792 \\
\bottomrule
\end{tabular}
}
\caption{\textbf{Image Editing Results.} Comparison of different methods on the MagicBrush~\cite{zhang2024magicbrush} testset.}
\vspace{-4mm}
\label{tab:image_edit}
\end{table}

\begin{table*}[t]
\centering
\scalebox{0.85}{
\begin{tabular}{lccccccccc}
\toprule
\multirow{3}{*}{Settings} & \multicolumn{2}{c}{Low Light Enhancement} & \multicolumn{2}{c}{Derain} & \multicolumn{2}{c}{Object Addition} & \multicolumn{2}{c}{Object Removal} & Depth Estimation\\
 & \multicolumn{2}{c}{LOL} & \multicolumn{2}{c}{Rain100H} & \multicolumn{4}{c}{InstructP2P~\cite{brooks2023instructpix2pix}} & NYU-v2 \\
         & PSNR↑ & SSIM↑ & PSNR↑ & SSIM↑ & $\text{CLIP}_{\text{dir}}$↑ & $\text{CLIP}_{\text{out}}$↑ & $\text{CLIP}_{\text{dir}}$↑ & $\text{CLIP}_{\text{out}}$↑ & RMSE↓ \\
\midrule
OmniGen~\cite{xiao2024omnigen} (in-context) & 8.923 & 0.243 & 13.14 & 0.411 & 0.054 & 0.243 & 0.031 & 0.233 & $\times$\\
\midrule
Full training & \textcolor{gray!70}{19.71} & \textcolor{gray!70}{0.770} & \textcolor{gray!70}{21.55} & \textcolor{gray!70}{0.633} & \textcolor{gray!70}{0.112} & \textcolor{gray!70}{0.283} & \textcolor{gray!70}{0.103} & \textcolor{gray!70}{0.265} & \textcolor{gray!70}{0.279} \\
No In-context & 9.140 & 0.253 & 7.924 & 0.212 & -0.031 & 0.252 & 0.023 & 0.244 & 0.745 \\

In-context w/o \methodname & 17.00 & 0.633 & 18.10 & 0.509 & 0.092 & 0.262 & 0.069 & 0.246 & 0.390 \\
In-context w/ \methodname & 17.22 & 0.653 & 18.91 & 0.512 & 0.092 & 0.274 & 0.073 & 0.251 & 0.352 \\

\bottomrule
\end{tabular}
}
\vspace{-3mm}
\caption{\textbf{Results of in-context learning in novel task settings.} ``Full training'' denotes for model trained with corresponding training set. While the other settings evaluate performance on tasks not encountered during training.}
\label{tab:novel_task}
\end{table*}

\begin{figure*}[t]
\centering
\includegraphics[width=1.0\textwidth]{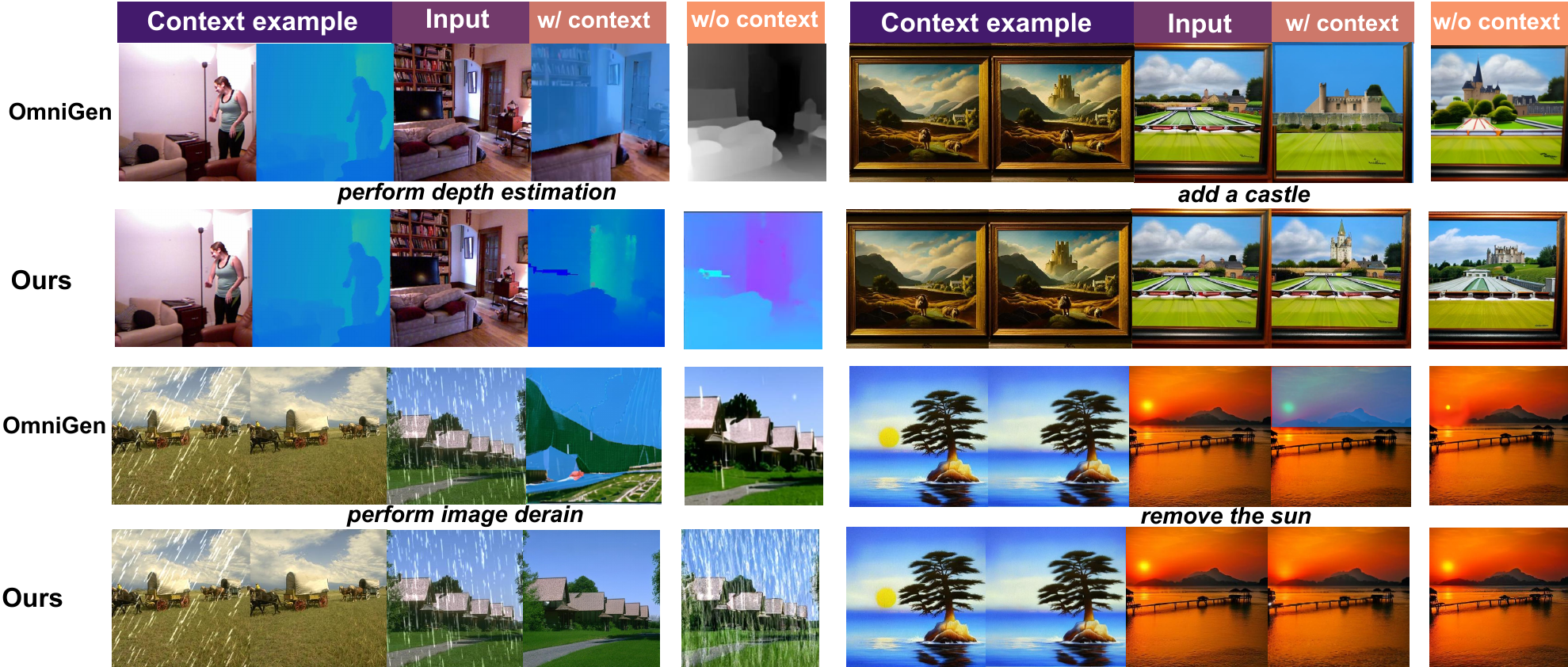}
\vspace{-4mm}
\caption{\textbf{Novel task in-context testing compared to OmniGen~\cite{xiao2024omnigen}.} \methodname can achieve novel task generalization with a given example. While OmniGen~\cite{xiao2024omnigen} fall short in in-context learning (such as adapting to new color spectrum or preserve details when adding object to the image).
}
\vspace{-3mm}
\label{fig:com_omnigen}
\end{figure*}

\noindent \textbf{Results.} 
As shown in \cref{tab:image_edit}, training solely with image editing pairs does not yield satisfactory results, as the model tends to replicate the original image rather than apply meaningful edits. Incorporating the additional difference description task through training model with next text token prediction encourages the model to identify distinctions between the input and edited images. This significantly improve the overall performance. Testing with an editing pair retrieved from the training set serving as an in-context example further enhances the quality of the edits, which proves the effectiveness of RAIE. Qualitative results are presented in \cref{fig:edit}. This experiment demonstrates the effectiveness of our in-context learning strategy for image editing tasks that require advanced semantic comprehension.

\subsection{In-Context Learning on Novel Tasks}

\noindent \textbf{Settings.} 
Following the approach of GPT-3~\cite{mann2020language}, this experiment primarily investigates the generalizability of our model on novel tasks, given only a single example as context. We choose Low light enhancement and Image derain from low-level vision, object addition and object removal from Image Editing as novel task to perform this study. During training, we remove the training data for each novel tasks. For Image derain, we remove the training of generating derained image of Rain-13K. For low light enhancement, we remove the training of generating enhanced image of Mit-5K~\cite{fivek} and LOL~\cite{wei2018deep}. For image editing task. we filtered out the sample in Ultra-Edit~\cite{zhao2024ultraedit} and MagicBrush~\cite{zhang2024magicbrush} using LLama-3-Instruct-7B~\cite{dubey2024llama} by querying whether the instruction involves object addition or removal. Our test is perform on Rain100H~\cite{yang2017deep} (Image derain), LOL-val (low light enhensement) and manually filtered 100 editing samples of Object Addition/ Removal task from publicly available instructP2P~\cite{brooks2023instructpix2pix} dataset. For depth estimation we test a new color palette that match the distance of the pixel on image to a different color spectrum. For low level vision, we report PSNR and SSIM as image evaluation metric. For Image Editing, we report CLIP~\cite{radford2021learning} score consistent with ~\cite{zhao2024ultraedit,zhang2024magicbrush}. For depth estimation task, we report the RMSE on the previous unseen color spectrum.

\begin{figure*}[t]
\centering
\includegraphics[width=1.0\textwidth]{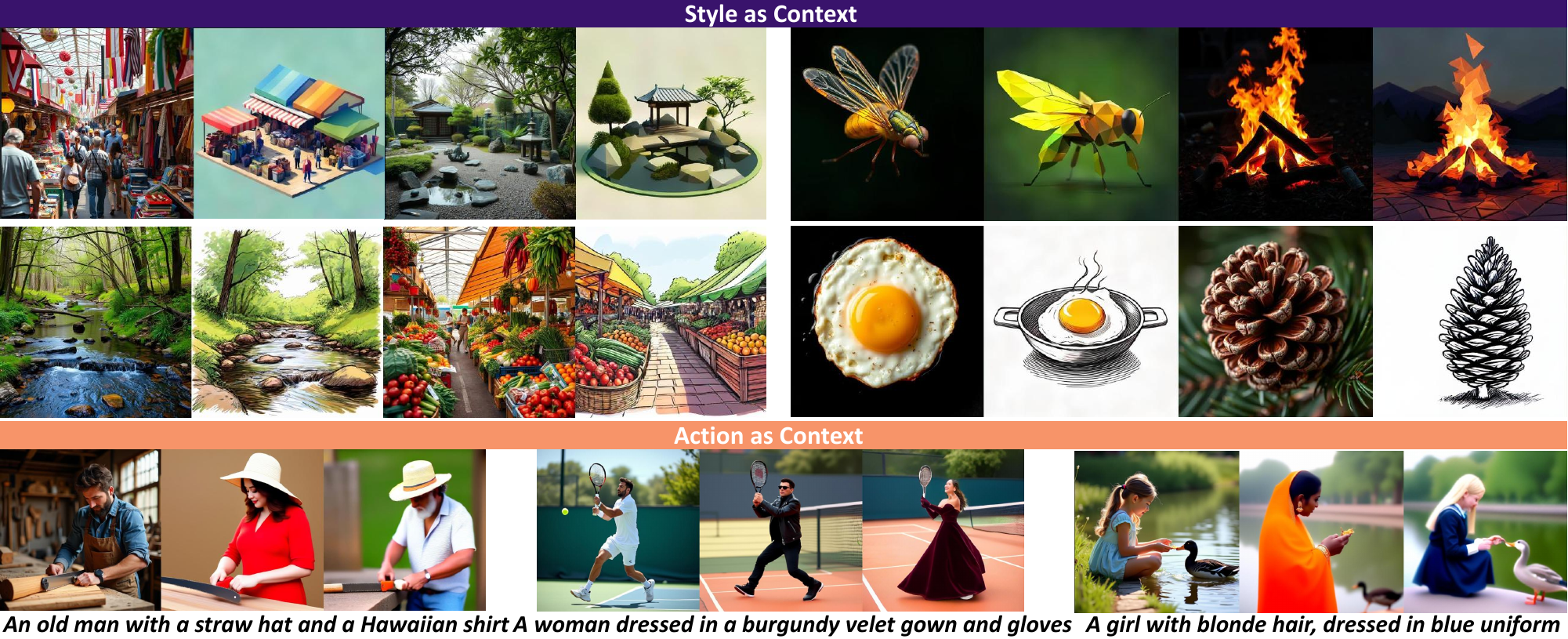}
\vspace{-4mm}
\caption{\textbf{\methodname can support diversified context} to achieve style personalization and action preservation.
\vspace{-3mm}
}
\label{fig:style_and_action}
\end{figure*}

\noindent \textbf{Results.} 
We report the quantitative results in \cref{tab:novel_task}. ``Full training" refers to the model fully trained to corresponding training set. Both ``In-context" and ``No In-context" settings are trained with the corresponding training data completely removed. On novel task, in-context example can significantly improve results. In contrast, the model generally failed to perform the task without the in-context example. We also ablate the effectiveness of attention masking proposed in ~\cref{sec:attn_mask}, where \methodname's attention mask force the model compresses the in-context information implicitly. This achieves slightly improved result compared to standard causal masking.

We also compare our in-context learning capabilities with OmniGen, where we input three image follow the same prompt template in ~\cite{xiao2024omnigen}. As shown in both \cref{tab:novel_task} and \cref{fig:com_omnigen}, due to prohibitive context-length (with the image resolution already reduced to 512x512), OmniGen struggled to achieve generalized in-context learning. In depth estimation, it fails to generalize to unseen color spectrums. In image editing, OmniGen is unable to keep unchanged parts of the image consistent with the original, nor did it effectively follow contextual cues. For image deraining, the model struggled to interpret the context accurately, leading to unexpected results. In contrast, our model, leveraging unified text and image next-token prediction loss, demonstrates superior generalization to previously unseen tasks.

\subsection{Other In-context Form}

\noindent \textbf{Settings.} 
In addition to training our model on existing datasets, we also create two small datasets for style personalization and action preservation to demonstrate \methodname's ability to extract diverse contextual information. For style-personalization, we use RB-Modulation~\cite{rout2024rb} to generate image pairs based on style image and further filter low quality data with QWen-VL2~\cite{wang2024qwen2} (detailed in \cref{sup:rb_modulation}). For action preservation, we generate diversified human actions and use pose estimation model and ControlNet~\cite{zhang2023adding} to generate different person doing same action in the same pose. For each task, we generate 10K pairs for in-context learning. For style personalization, we give model an example transformation pair to prompt model perform similar transformation on an unseen image. For action and pose preservation, we give model two image of a person doing same action in similar pose and a new person description and prompt model to generate a new image.

\noindent \textbf{Results.} We show qualitative results in \cref{fig:style_and_action}. \methodname can extract both the high level semantics and low level details of the context example and perform successful transformation on new image or generation based on text prompt. This experiments demonstrate that \methodname can achieve diversified in-context form in arbitrary multi-modal tasks.

%% file: sec/5_discussion.tex
\section{Discussion}
In this work, we propose empowering the autoregressive foundation model Chameleon~\cite{team2024chameleon} to achieve unified image generation through in-context learning. We demonstrate its promising performance across tasks such as text-to-image generation, dense prediction, low-level vision, and image editing, and showcase its generalizability to previously unseen tasks when provided with in-context examples. We hope this work will pave way for this promising direction to achieve the ``GPT-3 moment" in the unified multi-modal field in image generation.

While promising, our work still faces several unresolved challenges. First, the VQ-VAE in our base model, Chameleon~\cite{team2024chameleon}, introduces substantial information loss in image reconstruction at a compression rate of 16, which is a primary reason for the model’s reduced performance on certain low-level vision tasks that demand high-quality image reconstruction. Second, \methodname achieves successful in-context generalization only when a sub-task within a prototype task (as shown in \cref{fig:data}) is excluded from training, with limited generalization across different prototype tasks. We believe that more comprehensive and diversified multi-modal pretraining is essential to bridge the gaps between different prototype tasks, ultimately achieve the ``GPT-3 moment" in multi-modal learning.

\section{Acknowledgment}


\vspace{-1.5mm}
This project is funded in part by Shanghai Artificial lntelligence Laboratory, the National Key R\&D Program of China (2022ZD0160201), the Centre for Perceptual and Interactive Intelligence (CPII) Ltd under the Innovation and Technology Commission (ITC)’s InnoHK. Dahua Lin is a PI of CPII under the InnoHK.

%% file: sec/X_suppl.tex
\clearpage
\appendix
\setcounter{page}{1}

\twocolumn[{%
\renewcommand\twocolumn[1][]{#1}%
\maketitlesupplementary

\begin{center}
    \vspace{-10pt}
    \centering
    \captionsetup{type=figure}
    \includegraphics[width=1.0\textwidth]{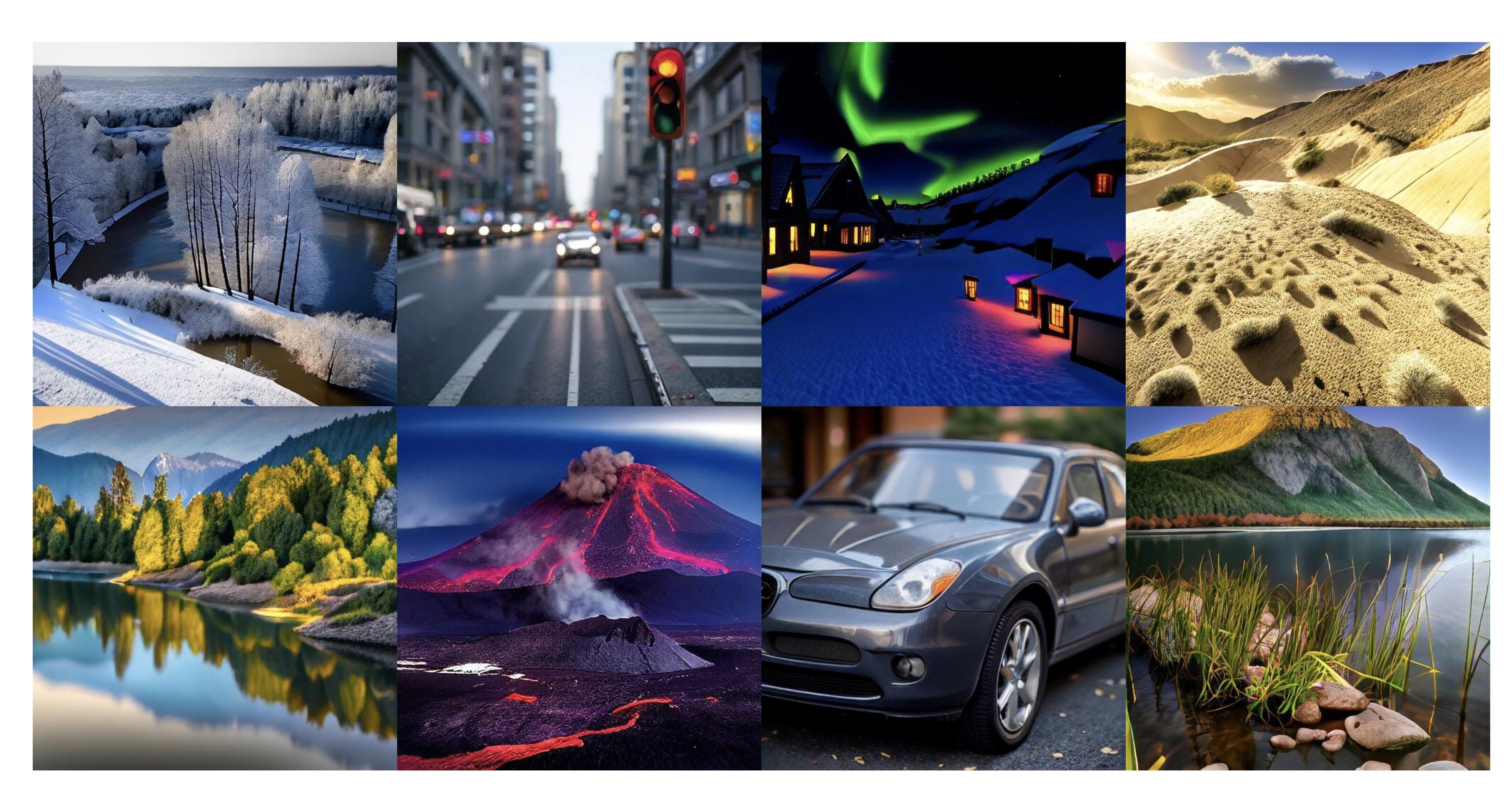}
    \vspace{-10mm}
    \caption{\textbf{Qualitative results of text-to-image generation.} High-quality text-to-image generation cases with high aesthetic qualities after training on Laion-Aesthetics~\cite{schuhmann2022laion}.}
    \label{fig:T2I1}
    \vspace{2mm}
\end{center}
}]


\section{Qualitative Results of Text-to-Image Generation.}
\label{sup:t2i}
\begin{figure*}[t]
\centering
\includegraphics[width=1.0\textwidth]{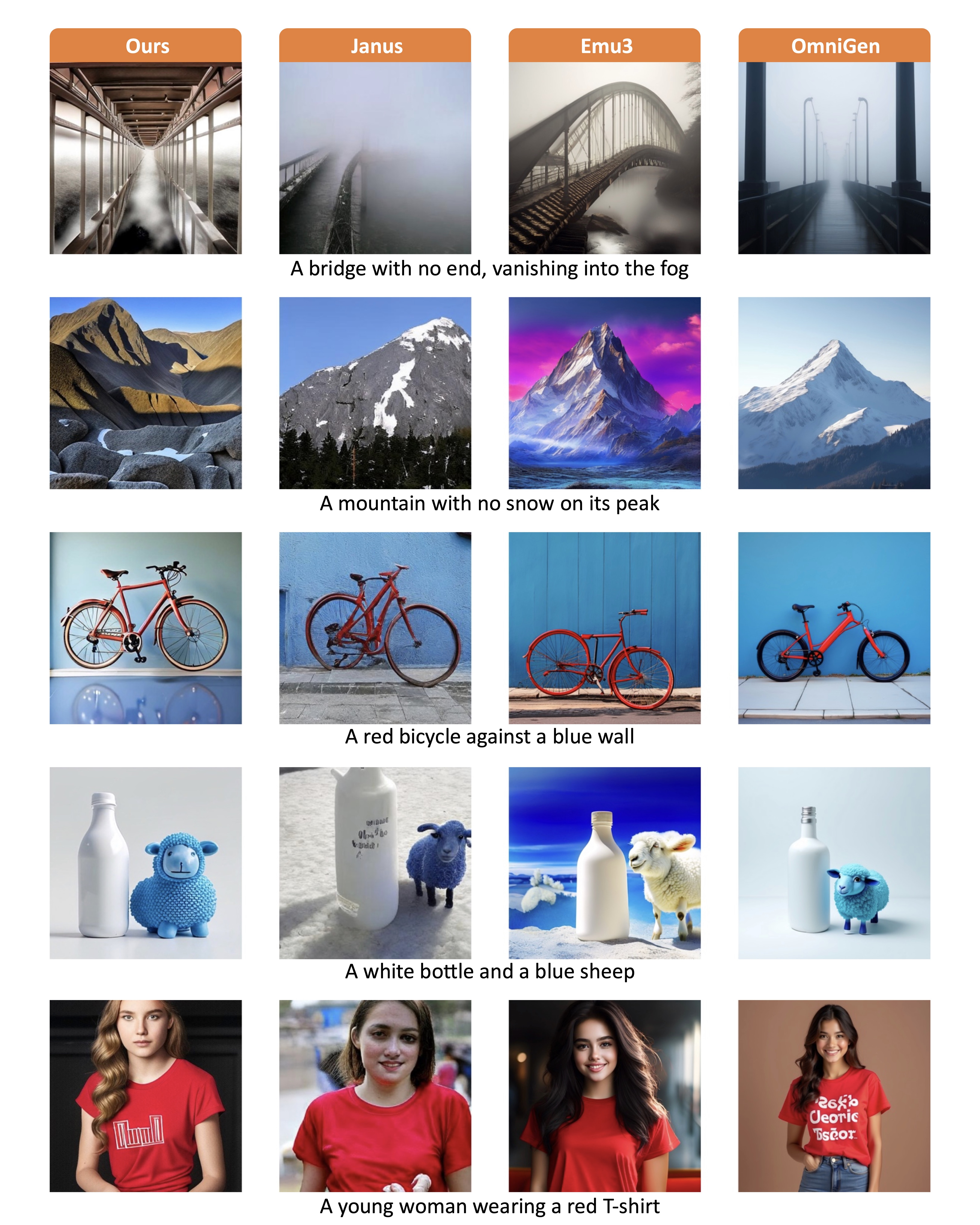}
\caption{\textbf{Qualitative results of text-to-image generation.} Compared to Janus~\cite{wu2024janus} and Emu3~\cite{wang2024emu3}, our model presents marked improvement in both quality and textual alignment.}
\label{fig:T2I2}
\vspace{-2mm}
\end{figure*}

We visualize some text-to-image generation results of our model in \cref{fig:T2I1} and comparison with other models in \cref{fig:T2I2}. \cref{fig:T2I1} demonstrates that our model can generate images with high aesthetic qualities after training on filtered high qaulity data from Laion-Aesthetics~\cite{schuhmann2022laion}. \Cref{fig:T2I2} clearly demonstrates that the incorporation image dense description task has significantly bolstered the model’s proficiency in accurately following text prompt when compared to other models such as Emu3~\cite{wang2024emu3} and Janus~\cite{wu2024janus}.

\section{Details of training data}
\label{sup:data}

Full training data statistics are reported in \cref{tab:train_aug} and \cref{tab:train_no_aug}. For each of the task in \cref{tab:train_aug}, we use QWenVL-2~\cite{wang2024qwen2} to describe the transformation between the input and output images and augments with reversed task.

\section{Details of Prompt template}
\subsection{Difference description task.}
\label{sup:diff_description}
For each image editing pair in Ultra-Edit~\cite{zhao2024ultraedit} and MagicBrush~\cite{zhang2024magicbrush}, we leverage QWenVL2~\cite{wang2024qwen2} to describe the difference between images using the following prompt: ``Describe the differeces between two images. Use `input image' describe the first image and `output image' to describe the second image, describe what subtask it belongs to, choosing from [Style Transfer, Object Removal, Object Replacement, Object Addition, Object Modification, Season/ Time Change, OTHER\_SUBTASK]". We also ask QWenVL2 to label a reverse editing prompt for data augmentation.

\subsection{filtering data generated by RB-Modulation.}
\label{sup:rb_modulation}

To generate and filter high-quality data, we first use FLUX to generate high-quality and stylized images based on the prompt templates in \cref{tab:style}. However, RB-Modulation~\cite{rout2024rb} occasionally performs correct style transformations but sometimes fails. To ensure quality, we further use QWen-VL2~\cite{wang2024qwen2} for data filtering. Due to QWen-VL2's current limitations in analyzing relationships among three images, we conduct quality filtering in two stages. First, we ask QWenVL-2 to verify the consistency of the main object and semantic with the base image using the question: ``Do you think the two image shares the same semantics and basic layout [Yes/No]? Provide your reasoning.". Next, we check the success of style transfer from exemplar image by asking ``Do you think the two image shares the same style [Yes/No]? Provide your reasoning." Through this process, we filter 10K high quality image based style-personalization image pairs to incorporate into the training of \methodname.

\subsection{filtering data generated by IP-Adapter.}

\begin{figure*}[t]
\centering
\includegraphics[width=1.0\textwidth]{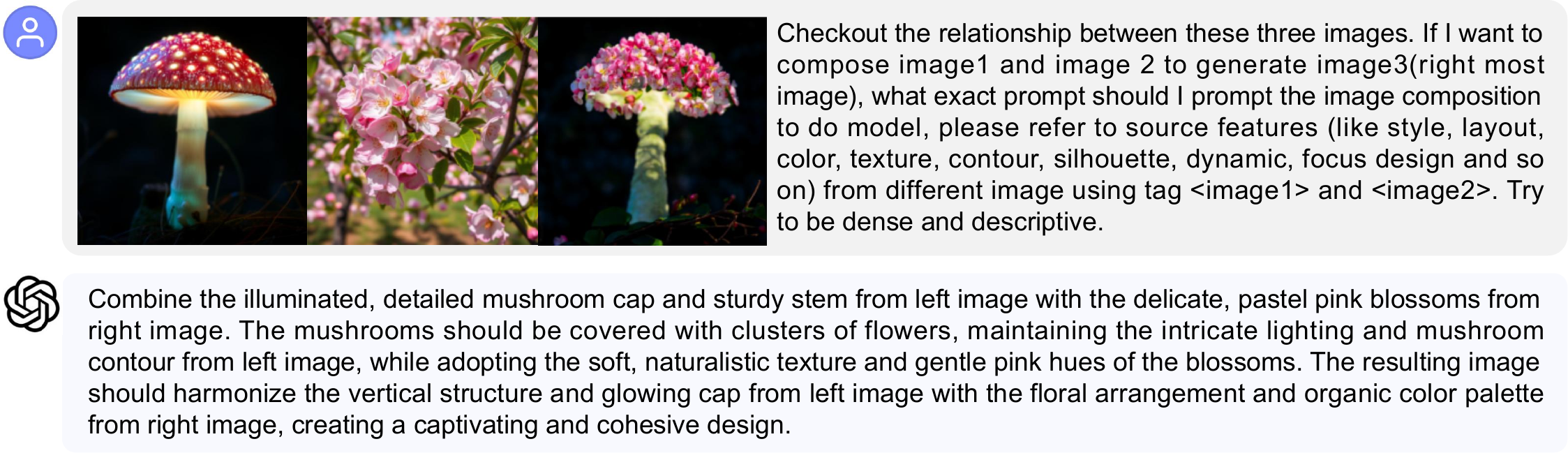}
\caption{\textbf{An example of conversation with GPT-4o} to annotate the relationship between input images and output image produced by IP-Adapter~\cite{ye2023ip}}
\label{fig:ip_adapter}
\vspace{-2mm}
\end{figure*}

IP-Adapter~\cite{ye2023ip} can perform layout and semantic combination on two provided images, However, the final output image can maintain different attributes (layout, semantics, texture, details) from different images in a unified but not entirely deterministic format. Given the complex attributes relationship between the input images and the output images, we employ GPT-4o to analyze and annotate these relationships. As shown in \cref{fig:ip_adapter}, GPT-4o provides high-quality, detailed descriptions of the relationships between different images. For this purpose, we annotate a dataset of 50K image pairs.

\subsection{Caption Rewriting on Laion-aesthetic.}
\label{sup:cap_laion}
We filter high-quality data from Laion-Aesthetic~\cite{schuhmann2022laion}, selecting images with an aesthetic score greater than 6. For dense caption rewriting, we use QWen-VL2~\cite{wang2024qwen2}, focusing on the relative positions, colors, and numbers of objects. To preserve caption diversity, we retain 10\% of the original captions during training.

\section{Retrieval-Augmented Image Editing}
\label{sup:raie}
\begin{figure*}[t]
\centering
\includegraphics[width=1.0\textwidth]{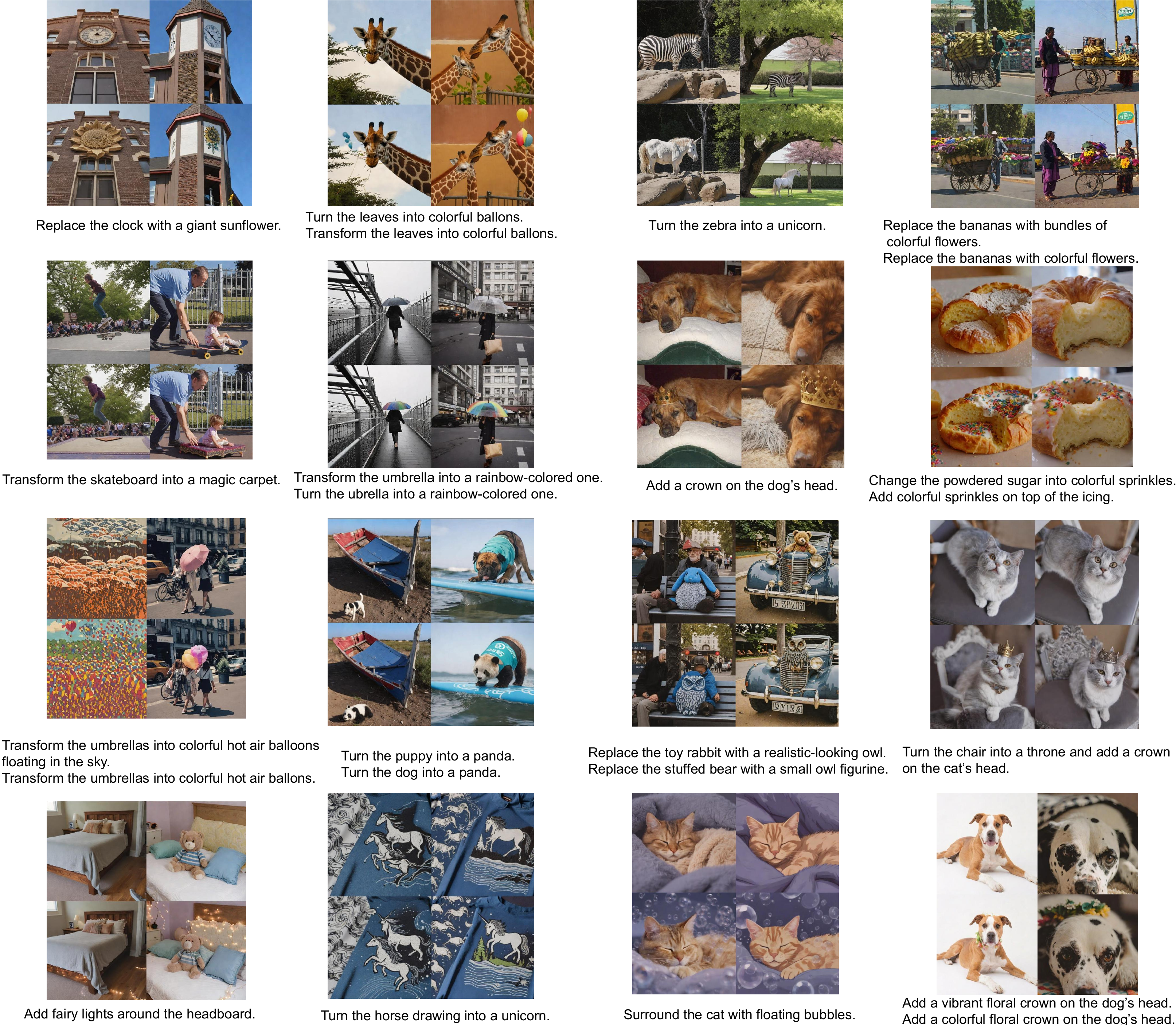}
\caption{\textbf{Visualization of in-context training example in RAIE.} After CLIP~\cite{radford2021learning} based clustering, many instruction are similar or completely the same, which is crucial to the success of RAIE.}
\label{fig:raie}
\vspace{-2mm}
\end{figure*}

Clustering similar editing pairs during training is critical to the success of Retrieval-Augmented Image Editing (RAIE) as a form of in-context learning. Fortunately, we observe that many editing instructions in MagicBrush~\cite{zhang2024magicbrush} and UltraEdit~\cite{zhao2024ultraedit} are highly similar to each other. As shown in \cref{fig:raie}, by pairing each editing pair with its nearest neighbor based on CLIP~\cite{radford2021learning} text feature similarity, we find that many instructions are either similar or identical. This similarity is a key factor contributing to the effectiveness of RAIE.

\section{More Qualitative Results Visualization.}
\begin{figure*}[t]
\centering
\includegraphics[width=1.0\textwidth]{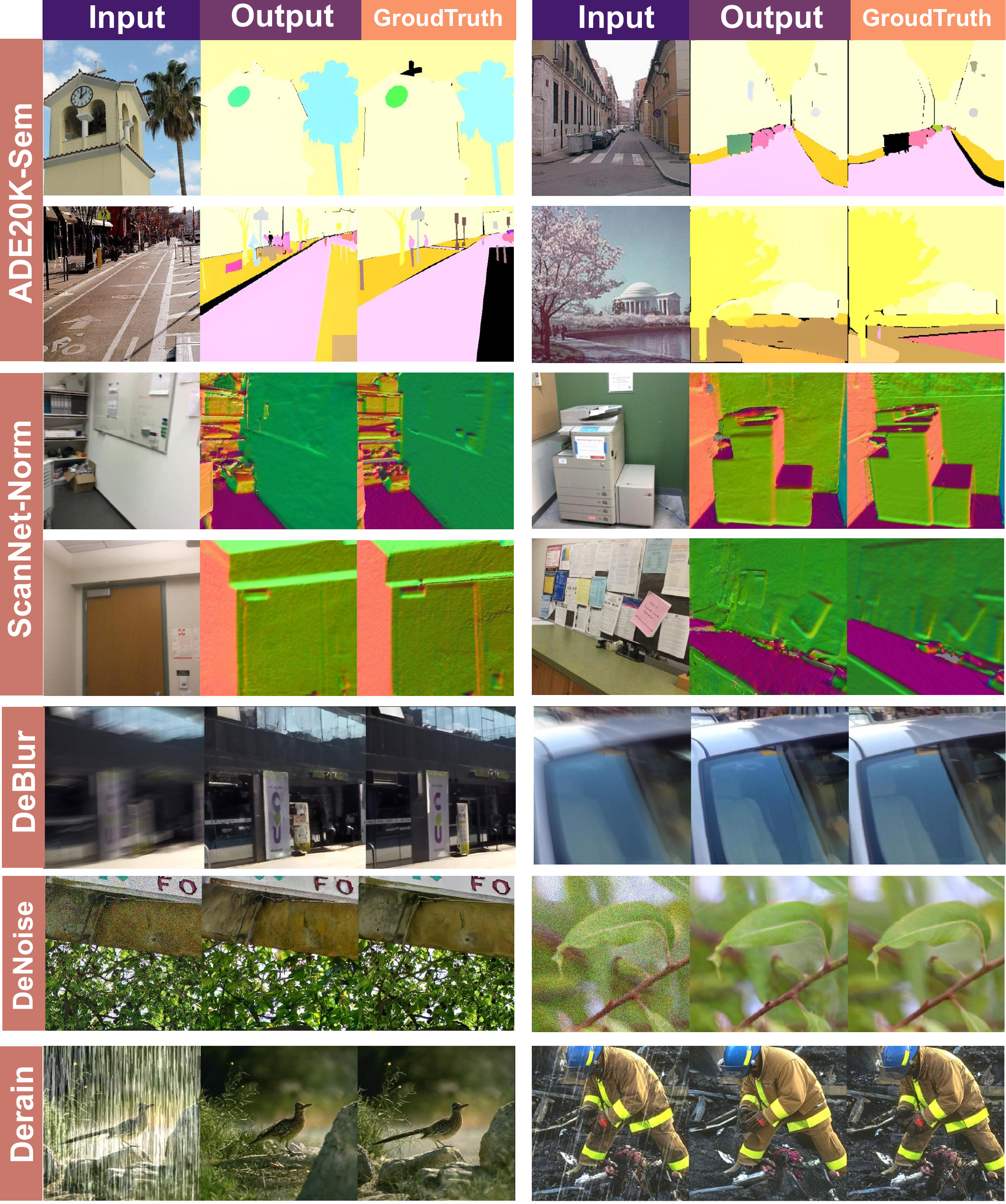}
\caption{\textbf{Qualitative results of diversed tasks,} such as semantic segmentation, norm estimation, image deblur, denoise and derain.}
\label{fig:tasks}
\vspace{-2mm}
\end{figure*}

We provide more visualization on vision tasks in \cref{fig:tasks}.

\newpage
\section{Higher Resolution Reconstruction}
\begin{table}[h]
  \centering
  \scalebox{0.95}{
    \begin{tabular}{lccc}
    \toprule
    Model & Resolution & PSNR  & SSIM \\
    \midrule
    \multirow{2}[0]{*}{SDXL-VAE (16x)} & 512   & 27.51 & 0.810 \\
          & 1024  & 32.13 & 0.922 \\
    \midrule
    \multirow{2}[0]{*}{Chemeleon-VQVAE (16x)} & 512   & 26.34 & 0.805 \\
          & 1024  & 29.77 & 0.906 \\
    \midrule
    Emu3-VQVAE (8x) & 512   & 27.78 & 0.833 \\
    \bottomrule
    \end{tabular}%
    }
  \caption{\textbf{Reconstruction quality tested on Rain-100L}. Increasing resolution can greatly enhance reconstruction quality.}
  \label{tab:reconstruct}%
\end{table}%

We use the Rain-100L derained test set to evaluate the reconstruction abilities of different models. As shown in \cref{tab:reconstruct}, increasing the input resolution significantly enhances reconstruction quality for both VQ-VAE and VAE models. This improvement arises from the fact that image compression inherently leads to a loss of detail, and providing higher-resolution input allows the model to recover previously lost details, resulting in better outputs. However, we are unable to implement \methodname with a 1024 resolution as Chameleon~\cite{team2024chameleon} is pretrained exclusively on a 512 resolution. We anticipate significant improvements across all tasks with the availability of higher-resolution early-fusion multi-modal foundation models in the future.

\begin{table*}[htbp]
\setlength{\tabcolsep}{2pt}  
  \centering
  \scalebox{0.75}{
    \begin{tabular}{lcccccccccccc}
    \toprule
          & DFWB  & GoPro & Rain13k & mit5k & LoL   & Laion\_Aesthetic & Ultra-Edit & MagicBrush & NYU-v2-depth & ADE20K & ScannNet-Norm & dep/seg/norm/hed/mlsd2img \\
    \midrule
    Ori\_data & 72K   & 17K   & 13K   & 5K    & 6K    & 500K  & 500K  & 1.7K  & 48K   & 20K   & 260K  & 100K $\times$ 5 \\
    Augmentation & 288K  & 68K   & 52K   & 20K   & 24K   & 1000K & 2000K & 6.8K  & 192K  & 80K   & 1040K & 100K $\times$ 20 \\
    \bottomrule
    \end{tabular}%
    }
    \caption{\textbf{Detailed statistics of training data with augmentation.} For each pair, we use reverse task and difference description task to augment the data.}
  \label{tab:train_aug}%
\end{table*}%

\begin{table*}[htbp]
  \centering
    \begin{tabular}{lccccc}
    \toprule
          & RB-Modulation & IP-Adapter & Viton-Try-On & Pose\&Action & MimicBrush \\
    \midrule
    Ori\_data & 10K   & 50K   & 120K  & 10K   & 50K \\
    \bottomrule
    \end{tabular}%
    \caption{\textbf{Detailed statistics of training data without augmentation.}}
  \label{tab:train_no_aug}%
\end{table*}%

\begin{table*}[htbp]
\begin{tabular}{>{\raggedright\arraybackslash}p{0.3\textwidth} 
                >{\raggedright\arraybackslash}p{0.3\textwidth} 
                >{\raggedright\arraybackslash}p{0.3\textwidth}}
\toprule
Rendering           & Sketch             & Poster             \\
Mosaic              & Cyberpunk          & Line Drawing       \\
Impressionism       & Cubism             & Surrealism         \\
Pop Art             & Minimalism         & Abstract Expressionism \\
Pointillism         & Graffiti           & Neo-Futurism       \\
Baroque             & Melting Gold       & Melting Silver     \\
Gold Material       & Made by Gold       & Made by Bronze     \\
Fantasy Glow Rendering & Low-Poly        &                    \\
\bottomrule
\end{tabular}
\caption{\textbf{Style key words} for FLUX to generate stylised images.}
\label{tab:style}%
\end{table*}%

